\documentclass{article}

\usepackage{PRIMEarxiv}

\usepackage[utf8]{inputenc} % allow utf-8 input
\usepackage[T1]{fontenc}    % use 8-bit T1 fonts
\usepackage{hyperref}       % hyperlinks
\usepackage{url}            % simple URL typesetting
\usepackage{booktabs}       % professional-quality tables
\usepackage{nicefrac}       % compact symbols for 1/2, etc.
\usepackage{microtype}      % microtypography
\usepackage{lipsum}
\usepackage{fancyhdr}       % header

\usepackage{bbm}
\usepackage{amsmath}
\usepackage{graphicx}
\usepackage{algorithm, algorithmic}
\usepackage{amsfonts} 

\title{Heterogeneous Federated Reinforcement Learning Using
Wasserstein Barycenters

}

\author{
  Luiz Manella Pereira, M. Hadi Amini \\
  Knight Foundation School of Computing and Information Sciences \\
  Florida International University \\
  Miami\\
  \texttt{\{lpere339, moamini\}email@email} \\
}

\begin{document}
\maketitle

\begin{abstract}
In this paper, we first propose a novel algorithm for model fusion that leverages Wasserstein barycenters in training a global Deep Neural Network (DNN) in a distributed architecture. To this end, we divide the dataset into equal parts that are fed to "agents" who have identical deep neural networks and train only over the dataset fed to them (known as the local dataset). After some training iterations, we perform an aggregation step where we combine the weight parameters of all neural networks using Wasserstein barycenters. These steps form the proposed algorithm referred to as FedWB. Moreover, we leverage the processes created in the first part of the paper to develop an algorithm to tackle Heterogeneous Federated Reinforcement Learning (HFRL). Our test experiment is the CartPole toy problem, where we vary the lengths of the poles to create heterogeneous environments. We train a deep Q-Network (DQN) in each environment to learn to control each cart, while occasionally performing a global aggregation step to generalize the local models; the end outcome is a global DQN that functions across all environments.
\end{abstract}

% keywords can be removed
\keywords{Optimal Transport \and Wasserstein Barycenters \and Reinforcement Learning \and Federated Learning}

\section{Introduction}\label{introduction}
Reinforcement learning (RL) is among the core domains of machine learning (ML) and one of the pillars for real-world applications (e.g., autonomous driving, RLHF \cite{christiano2017deep}). Proper generalization of RL models requires robust and accurate representations of the environment an agent interacts with. To achieve this global model, we must serially train until convergence over each environment. Furthermore, it is often the case that virtual representations of real environments are not fully accurate and thus it is desirable to have a combination of training in virtual environments and in the real-world \cite{choo2020reinforcement}. The idea of combining both virtual and real environments leads us to the domain of federated learning. In Federated Reinforcement Learning (FRL), we have a homogeneous architecture, where each agent operates in the same environment, or a heterogeneous architecture, where each agent operates in a different environment \cite{qi2021federated}. In the related works section we will dive deeper into the existing solutions for the problem just laid out. Our goal is to introduce a novel approach leveraging Wasserstein barycenters to perform model fusion in heterogeneous federated reinforcement learning.

We begin by introducing the theory of model fusion and its functionality in deep learning. We follow it with an introduction to heterogeneous federated reinforcement learning. Following the Introduction is Related Works, section \ref{related_works}, where we introduce prior work for each topic. We follow it up with the Preliminaries Section (section \ref{preliminaries}),  where we lay out the base material for each topic pertaining to our paper. Section \ref{problem_formulation}, the Problem Formulation Section, describes in detail what problem(s) we are trying to solve and how we go about arriving at a solution. The Results section (\ref{results}) is broken into two subsections, one per experiment. The first subsection contains the design and goal of that experiment. In the second subsection, we present the results associated with each experiment. Lastly, we end our paper with our Conclusions (\ref{conclusion}), where we summarize the main idea and the results associated with it.

\subsection{Model Fusion in Deep Learning}
Deep learning has, for a long time, remained a serially dependent procedure where training must make iterative passes over an entire dataset. One can either consider the entire set as one batch (full-batch), split the data into smaller (equally sized) batches (mini-batch), or consider each data point as a separate batch (online or stochastic). As we iterate over an entire dataset, we call this a singular epoch. When training a deep neural network (DNN), we make multiple passes over the entire dataset, and thus have multiple epochs. With the emergence of very large models, we need to consider attentively the size of the data, the size of the batches, and the number of training epochs, as they can contribute to a time-consuming training process or require a large amount of memory. For example, we can consider OpenAI's GPT model series. Each model in the sequence of GPT models they have released has increased in size, from a few billion, \cite{solaiman2019release}, to tens of billions, to over a hundred billion parameters. Training large models relies on extremely large datasets, and processing these sets sequentially takes a long time. In this paper, we propose fusing models during the training process, allowing deep neural networks to parallelize training. We leverage Wasserstein barycenters to perform model fusion, with the intent of speeding up the processing time, as we can now cover batches concurrently rather than serially. We showcase the idea by creating a DNN to classify MNIST data under a distributed architecture.

\subsection{Heterogeneous Federated Reinforcement Learning}

\begin{figure}
    \centering
    \includegraphics[scale=0.45]{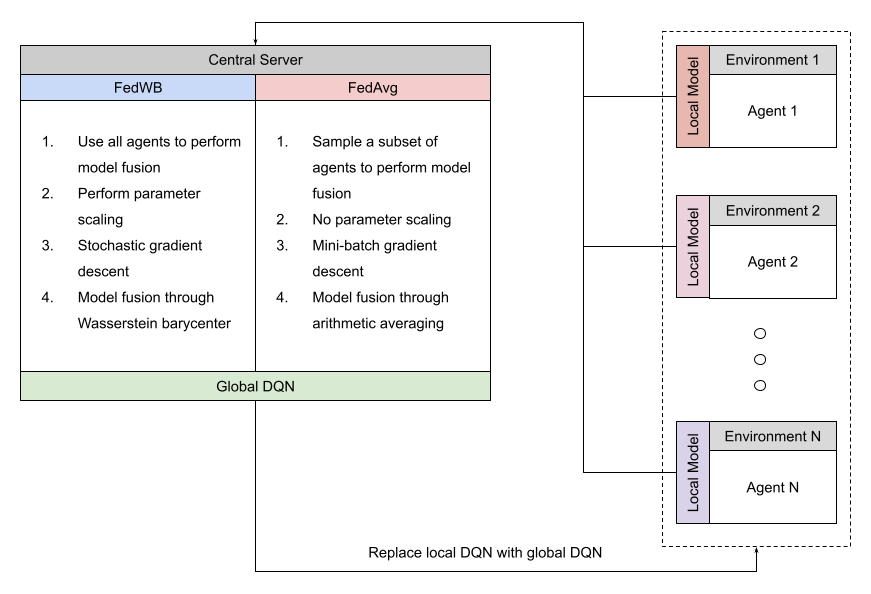}
    \caption{Heterogeneous Federated Reinforcement Learning (HFRL) - Model Architecture. Model architecture for HFRL. The table on the left places FedWB and FedAvg in juxtaposition. On the right-hand side, there are three boxes, each representing an agent, its local model, and its local environment. Pointers exist going from right to left to indicate the flow of local information from each agent, and one exists pointing from left to right to indicate the sharing of the global model that is broadcast back to the agents.}
    \label{fig:hfrl_model_architecture}
\end{figure}

One of the main components of reinforcement learning is the environment the agent interacts with. In many cases, the environment remains ``relatively'' static (e.g., in Go the board doesn't change \cite{silver2016mastering}), while in other applications, the environment may change drastically (e.g., self-driving cars learning driving mechanics for rainy weather vs. normal/sunny conditions \cite{ishihara2021multi}). There are different ways of handling this environment change, such as Federated Reinforcement Learning for heterogeneous environments. In this paper, we aim to leverage the work we lay out on model fusion using Wasserstein barycenters for deep learning to develop a singular, global model that is capable of performing tasks and making decisions across various environments. Unlike previous papers, which typically train an agent over each environment until it can achieve its task before moving on to the next environment, we instead choose to train agents to become experts in their own domains while intermittently aggregating all their acquired knowledge to generate one model with decision-making capabilities in those various scenarios. Previous work have similar ideas but utilizes the arithmetic averaging in their systems. We instead choose to use Wasserstein barycenters for model aggregation. A high-level overview of our work can be seen in Figure \ref{fig:hfrl_model_architecture}.

\section{Related Works}\label{related_works}
\subsection{Model Fusion in Deep Learning}
Model fusion in deep learning is a relatively new topic, and its complexities have yet to be fully explored. Some very early methods, which at the time did not utilize the term model fusion, are those of federated learning. For example, in \cite{mcmahan2017communication}, the authors introduced the Federated Averaging algorithm, which uses arithmetic averaging to perform an aggregation process in order to create a global model; their algorithm train local agents while intermittently grabbing the local agent's neural network weights and average them to create a global neural network. We believe that other aggregation methods exist that use less naive averaging techniques and showcase this in this work.

More recent work use the Wasserstein barycenter tool we propose to use, albeit in a different form. In \cite{singh2020model}, the authors propose the first method for model fusion via optimal transport. They presented a layer-wise fusion algorithm for neural networks. The authors first align the neurons across varying networks and then perform vanilla averaging to obtain a global model. Akash et al., in \cite{akash2022wasserstein}, expanded on the work just mentioned and showed how Wasserstein barycenters can be used to fuse various NN types together. First, they defined probability measures over each layer in the networks. Coupled with a weight vector, they continued to take the Wasserstein barycenter over these probability measures for all networks. Through experimentation over heterogeneous and homogeneous datasets, they demonstrated the predictive capabilities of Wasserstein barycenters in "one-shot" model fusion in DNN, RNN, and LSTM.

\subsection{Federated Reinforcement Learning}
Federated reinforcement learning (FRL) brings forth a natural utility in the world of machine learning due to the privacy constraint that exists in the real world \cite{yang2019federated}. For example, in healthcare, the HIPPA holds clear rules regarding the sharing of patient information, and thus holds privacy at a state of utmost importance. It is not always the case that we have all the data centralized in a server where learning can take place. It is more often the case that data is distributed amongst agents across a network, and that each agent's information is private and should not be transmitted, at least not without some obfuscation \cite{wei2020federated}. Furthermore, simulations seldom represent the real world perfectly, adding yet another layer of complexity in training models. Federated learning has introduced solutions to tackle these problems and has recently been introduced to reinforcement learning, as is seen below.

Based on the literature of FRL, we notice an underlying principle that goes by various names, such as the critic or the model aggregator. Furthermore, we have noticed the aggregation process relies on variations of the weighted averaging function. Federation is introduced in RL in one of a few ways, but one notices a similar architecture. First, the architecture of the problem is a privacy-constrained network containing distributed, local information that must be accessed to train a model. These local devices are referred to as agents. In order to train a global model capable of generalizing across the board, each agent holds its own model, although they all have identical architecture, whose model information is shared from time to time. The shared information is then aggregated in the central server, which then broadcasts the model parameters back to the agents for local updates. The steps described are repeated until convergence or some stopping criteria are met. We see very little deviation from the structure described and will explain it here.

For a heterogeneous case, where each environment is different (e.g., in autonomous driving, different environments could mean different weather conditions), in \cite{pmlr-v151-jin22a}, the authors initialize $n$ agents in heterogeneous environments. The models, either Q networks or policy networks, train independently of each other except for the synchronization step. During the synchronization step, model parameters are sent to a central server, which generates a "global" model by performing an averaging; this simple averaging concept was used in the two models: $Q_{Avg}$ and $P_{Avg}$. The following equations explain the \textit{global aggregation} step and broadcasting of the new model:
\begin{equation}
    \Bar{Q}_t(s,a) \leftarrow \frac{1}{n}\sum_{i=1}^n Q_t^i(s,a), \forall s,a;
\end{equation}

\begin{equation}
    Q_t^i(s,a) \leftarrow \Bar{Q}_t(s,a), \forall s,a,k
\end{equation}

The authors explored some theoretical analysis of the two model choices and experimentally backed their results through simple applications (e.g., CartPole, Acrobat, Hopper, and Half-cheetah). We see this averaging technique again in \cite{khodadadian2022federated}, where the authors not only considered Q-learning but also considered on-policy and off-policy TD. Although fundamentally nothing differs in the aggregation process, Khodadadian et al. answer the following important questions:
\begin{enumerate}
    \item Does introducing $N$ agents in FRL yield, linearly, an $N$-fold speedup
    \item How does the convergence speed and final error scale with synchronization frequency
    \item What is the optimal number of synchronization steps
\end{enumerate}
In \cite{lim2021federated}, the authors aim to improve the learning efficacy of RL by introducing a novel federation policy for multiple agents. The authors introduced two different federation policy procedures. The first policy is the sharing of gradients to improve learning speeds; this is done by taking a weighted average of the gradients

\begin{equation}
    \overline{g} = \sum_{n=1}^N w^n g^n_\theta, \quad w^n = \frac{PI^a_\theta}{\sum_{n=1}^N PI^n_\theta},
\end{equation}

\noindent where $PI^a_\theta$ is an agent's Performance Index, N in the number of workers, and $PI^n_\theta$ is an individual worker's PI. The second federation policy is also introduced to help accelerate learning by sharing a mature (a model that has undergone further training) model's parameters with other workers. The transfer of weights is done intelligently based on each worker's individual PI scores by adding a weight component
\begin{equation}
    w^T = \frac{PI_t}{TC}
\end{equation}

\noindent where $PI_t$ is the Performance Index at time t and TC is the condition to stop learning in each environment. Each actor's parameters, $theta$, is then updated as such:

\begin{equation}
    \theta = w^T \times \theta + (1-w^T)\times \overline{\theta}
\end{equation}

\noindent where $\overline{\theta}$ is the parameters of the mature model. With the two procedures explained above, the authors simulated their efficacy in simulated environments (OpenAI's Cartpole, MountainCar, and Acrobot) and a real example (Quanser's QUBE-Servo 2 - Rotary Inverted Pendulum). In both simulated and real environments, applying their federation policies reduced the number of required episodes to train the models to maturity.

In \cite{nie2021semi}, the authors build a multiple UAV enabled MEC model with computation task offloading and resource management powered by a privacy centered Multi-Agent Federated Reinforcement Learning model. Their goal was to minimize the sum power consumption of the UAV-MEC system with the proposed DRL model to solve the control problem while superimposing a Gaussian differential privacy technique to maximize privacy. The primary novelty in the paper is the existence of multiple agents interacting in the same environment. These agents interact to accomplish tasks but maintain the same global model. The global model is trained in a similar fashion to prior-mentioned papers, with the exception of the use of Gaussian differentials as an encryption mechanism, onto which the global model uses these encrypted messages as inputs to their Q-networks. The final, and most important result shown was the experimental difference between the MARL model, which had no privacy constraint by fully sharing all local data, and the MAFRL model, which maintained privacy via encryption. Both models had very similar results, concluding that maintaining privacy, at least under the circumstances described in the paper, had negligible impact in the capabilities of the model to solve the problem at hand.

Unlike prior work, where global models were established through the sharing of gradients or encrypted values through the Gaussian differential privacy method, in \cite{lee2020federated}, the authors averaged the weights of the neural networks directly. Given local weights $w_i$ for $i=1,\dots,N$, the shared global model is given by simply averaging the weights. The sharing and updates are made until some stopping point is met. The novel idea presented was able to manage the energy consumption of multiple smart homes with A/Cs, washing machines, and photovoltaic devices (i.e., solar panels).

\section{Preliminaries and Definitions}\label{preliminaries}
\subsection{Neural Networks}
\begin{figure}[htp!]
    \centering
    \includegraphics[scale=0.4]{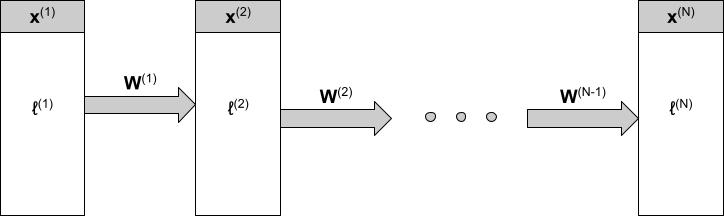}
    \caption{Compact NN design.}
    \label{fig:nn}
\end{figure}

Neural networks are the base model structures for deep learning \cite{lecun2015deep}. We can describe a fully connected neural networks by its components. Let $\ell_i$ be some layer $i$ of our network, given $i=1,\dots,N$, where $N$ is the total number of layers (including input and output layers). Let $\mathcal{W}^{(j)}$ be the weights used to forward propagate from layer $\ell_{j}$ to $\ell_{j+1}$ for $j=1,\dots,N-1$. As we forward propagate, we take the inputs, $x^{(j)}$, multiply them by $\mathcal{W}^{(j)}$, which is then fed into an activation function to generate the input to the next layer, $x^{(j+1)}=a(W^{(j)}x^{(j)})$. The output of the activation becomes the input to layer $\ell^{(j+1)}$. Figure \ref{fig:nn} shows the variables in the context of the network architecture but abstracts the number of nodes away into a compact representation where each layer has an arbitrary number of nodes with the expectation that the dimensions of $x$ and $W$ follow appropriately. Training of neural networks are done in traditional methods, using batch gradient descent \cite{khirirat2017mini} and backpropagation \cite{svozil1997introduction}.

\subsection{Preliminaries of Optimal Transport}
\subsubsection{General Theory}
Optimal transport (OT) deals with the idea of how to move mass from one location to another in a transportation-cost-minimizing fashion. While the original work was proposed by Monge \cite{monge1781memoire}, the more well-known formulation was proposed by Kantorovich \cite{kantorovich1942transfer}, where the idea of mass-splitting allowed for broadening of the applications of OT. The optimal transport problem is now formulated as follows: given that we have the following Monge maps,
\begin{equation}
    U(\mathbf{a},\mathbf{b}) = \bigg\{ P \in R_{+}^{n\times m} : P\mathbbm{1}_m = \mathbf{a} \quad \text{and} \quad P^T \mathbbm{1}_n = \mathbf{b} \bigg\},
\end{equation}
where $\mathbf{a,b}$ are probability vectors, the optimal transport problem can be written as:
\begin{equation}
    L_\mathbf{C}(\mathbf{a},\mathbf{b}) = \min_{P\in U(a,b)}\langle \mathbf{C},\mathbf{P} \rangle = \sum_{i,j} \mathbf{C}_{i,j}\mathbf{P}_{i,j},
\end{equation}
\noindent where $C$ is the cost matrix, $P$ is the permutation matrix defined by equation (6), and $i,j$ are the respective indexes of the matrices.

The extension of Kantorovich's work leads to the most important definition in OT, the p-Wasserstein distance. As defined in \cite{peyre2019computational}, the p-Wasserstein distance is
\begin{equation}
    W_p(\alpha,\beta)=\mathcal{L}_{d^p}(\alpha,\beta)^{1/p}
\end{equation}

\subsubsection{Wasserstein Barycenters}
Wasserstein barycenter is a form of averaging that retains information regarding the geometry of the underlying objects. It is computed as such:
\begin{equation}\label{wasserstein_barycenter}
    \min_{\mathbf{a}\in\Sigma_n}\sum_{s=1}^S\lambda_s W_p^p(\mathbf{a}, \mathbf{b}_s)
\end{equation}
In \ref{wasserstein_barycenter}, $\lambda_s$ is a weight parameter which is typically set to equally distributed. Unlike other averaging methods, where mass is placed according to each underlying object being averaged, Wasserstein barycenters focus on the retaining geometry. For example, in 1D unimodal probability distributions, traditional averaging yields a bimodal distribution \cite{eisenberger1964genesis}, with mass distributed underneath where each input distribution lies in space. Varying the weights varies the amount of mass placed under each input distribution. On the other hand, WBs yield a unimodal distribution that lies on a geodesic that connects the input distributions. Varying the weight parameter slides the output distribution along the geodesic closer to the input distribution whose weight is respectively higher.

\subsection{Reinforcement Learning}
In reinforcement learning (RL), all theories and models stem from dynamical systems, more specifically from optimal control theory in the form of the Markov Decision Process, also called an MDP \cite{li2017reinforcement}. MDPs are the classical formalization of sequential decision making, and are made up of a the following components: an agent, its environment, a policy which is used by the agent to make decisions, a reward, a value function, and lastly, although optional, a model of the environment. It is important to differentiate between what a reward is and what a value function is. The reward indicates what is good in the immediate, or short, term while the value function determines what is good in the long term; the interplay between these two determine the importance of the greedy decision versus the potential for higher value if the immediate-best choice is not played (delayed gratification). While training the agent, our goal is to maximize the expected return, which is defined as follows
\begin{equation}
    G_t = \sum_{k=0}^\infty \gamma^kR_{t+k+1}
\end{equation}
where $\gamma$ is the discount rate, $0 \leq \gamma \leq 1$, and $t$ is time. The discount rate determines the level of importance we place on the immediate rewards in comparison to the future rewards. Q-learning focuses on the action-valued function $Q^*: State \times Action \rightarrow \mathbb{R}$. The associated policy is defined by the max over the actions the agent can take given it is in state $s$:
\begin{equation}
    \pi^*(s) = \underset{a}{\text{argmax}}Q^*(s,a)
    \label{q-function}
\end{equation}
In deep Q-learning (DQL), we approximate the optimal Q-function given in equation (\ref{q-function}) using a neural network. The training update rules leverages the Bellman equation \cite{kirk2004optimal}
\begin{equation}
    Q^\pi (s,a) = r + \gamma Q^\pi(s', \pi(s'))
\end{equation}
Unique to RL, the commonly used loss function is the Huber loss function \cite{huber1992robust},
\begin{equation}
    \mathcal{L} = \frac{1}{|B|} \sum_{(s,a,s',r)\in B}\mathcal{L}(\delta)
\end{equation}
where $\delta = Q(s,a) - (r + \gamma \underset{a}{ \max} Q(s',a))$, $B$ is a batch, and 
\begin{equation}
    \mathcal{L}(\delta) = 
    \begin{cases}
        \frac{1}{2}\delta^2 & \text{for} |\delta| \leq 1 \\
        |\delta| - \frac{1}{2} & \text{otherwise}
    \end{cases}
\end{equation}

\section{Problem Formulation}\label{problem_formulation}
\subsection{FedWB - Model Fusion on MNIST}\label{model_fusion_problem_formulation_section}
Our goal is to tackle a federated, or distributed, learning architecture in deep learning using the novel theories of model fusion to parallelize training over agents. We begin by first creating the distributed architecture where a central server communicates and triggers processes over $D$ different agents. Each one of these agents receives an identical neural network, which is stored and maintained locally. We proceed by loading the MNIST dataset \cite{lecun1998gradient}, splitting it $n$ ways such that there is nearly equal representation across each division. With the architecture laid out, we now focus on training. Training works as follows, each agent will perform forward propagation and backpropagation over its data. It then takes a gradient step to update its local model weights. After this local epoch, the local server collects all local model weights to perform the model fusion step. Model fusion is described in algorithm \ref{model_fusion_algorithm}. There are two notable factors in the algorithm. First is the flattening of the matrix. By doing so we transform the WB problem into a 1D WB problem. Second, is the scaling factor. WB requires us to work with probability distributions, so we add the smallest value (to prevent negative numbers), flatten the matrix, and divide each term by the sum of the flattened matrix. We store the two scaling values to be averaged later and used to up-scale later. Model fusion outputs a global model which is distributed to the agents, who then update their local model with the new global model to continue training. After iterating over some number, $E$, of epochs, training is complete, and we evaluate the model using the respective test datasets. The steps just explained are described in algorithm \ref{nn_via_model_fusion}, which is the core work of this paper; we call this algorithm FedWB. Note that unlike FedAvg, FedWB utilizes stochastic gradient descent (SGD) instead of mini-batch gradient descent, since it aligns better with our work in HFRL. Another important technical point to compare is that FedAvg uses two hidden layers composed of 200 nodes each, whereas we use a single hidden layer with 256 nodes.

\begin{algorithm}
    \caption{Model Fusion for Deep Neural Networks using Wasserstein barycenters}
    \begin{algorithmic}[1]
        \REQUIRE $D$ $W_j^{(i)}$ weight matrix $i$ for agent $d \in D$ (D: number of agents in the analyzed network)
		\vspace{.07cm}
        \vspace{.07cm}
        % for each agent, take each of their weight matrices and normalize them
        % while storing the normalization factor
        \FOR{$d=1,\dots,D$}
		    \STATE $W_d^{(i)}$ := flatten($W_d^{(i)}$)
                \STATE $minimum\_weight_{id}$ = $\min \{W_d^{(i)}\}$
                \STATE $W_d^{(i)}$ := $W_d^{(i)}$ + $minimum\_weight_{id}$
                \STATE $scaling\_sum_{id}$ := $\sum_{k=1}^{|W_d^{(i)}|} (W_d^{(i)})_k$ 
                \STATE $W_d^{(i)}$ := $W_d^{(i)}$ / $scaling\_sum_{id}$
	    \ENDFOR
        % Compute the barycenters
	    \FOR{$i=1,\dots,N-1$}
		    \STATE $W^{(i)} := WB(W_1^{(i)}, \dots, W_D^{(i)}) * \frac{1}{D}\sum_{d=1}^D scaling\_sum_{id} - \frac{1}{D}\sum_{d=1}^D minimum\_weight_{id}$
            \STATE $W^{(i)}$ := reshape($W^{(i)}$)
        \ENDFOR
    \end{algorithmic}
    \label{model_fusion_algorithm}
\end{algorithm}

\begin{algorithm}[ht]
	\caption{FedWB}
	\begin{algorithmic}[1]
		\REQUIRE  All agents have their data stored and split into train/test, and have their model initiated 
        \REQUIRE Number of agents: $D$
        \REQUIRE Number of epochs: $E$
        \REQUIRE Number of batches for local training: $B$
		\vspace{.07cm}
        \vspace{.07cm}
        % for each agent, take each of their weight matrices and normalize them
        % while storing the normalization factor
		\FOR{$e=1,\dots,E$}
            \STATE Multithread inner loop
            \FOR{$d=1,\dots,D$}
                \FOR{$b=1, \dots, B$}
                    \STATE Train local model
                \ENDFOR
            \ENDFOR
            \STATE Aggregate agents' NN weights in server
            \STATE Create global NN by performing model fusion (algorithm \ref{model_fusion_algorithm})
            \STATE Broadcast global model and update local models
        \ENDFOR
        % Compute the barycenters
        \STATE Evaluate global model through predictions over local data
  \end{algorithmic}
	\label{nn_via_model_fusion}
\end{algorithm}

In order to properly study the results of our algorithm, we will perform various experiments to try and empirically show optimal choices for variables present. We consider the importance of the number of agents in our network and how the number of agents affects the convergence of a global solution. We further compare the amount of time it takes to train a global model up to a minimum of 90\% accuracy as we vary the number of agents in the network. In the Results section, we will graph and explain the results, and compare the results of FedWB against FedAvg and traditional, single agent, training of a NN.

\subsection{HFRL with Wasserstein Barycenters}
Leveraging the concepts and model presented in section \ref{model_fusion_problem_formulation_section}, we now extend the use of algorithm \ref{model_fusion_algorithm} and \ref{nn_via_model_fusion} to reinforcement learning; their combination naturally lead to a solution for a federated Q-network architecture as each Q-network is a neural network. By considering each NN in algorithm \ref{nn_via_model_fusion} as a Q-network, we convert the original formulation to the RL domain, and thus expand the application of FedWB towards HFRL, where underlying these problems is the utilization of WBs as a global aggregator. 

Given a heterogeneous architecture, where the environment is different for each node in our distributed network, we aim to train a global Q-network that is capable of handling the control problem of balancing a pole on a cart (the Cart-Pole problem). We begin by instantiating the environments with varying lengths of poles, hence creating heterogeneous environments. We create a target and online Q-network, each of which starts with the same set of parameters. Next, we distribute copies of these two networks to each agent to begin their training. Following the traditional training of DQNs, we train the online network, using the target network as the ground truth we try to predict towards. By traditional training we imply a similar structure as in Algorithm 1 of \cite{mnih2015human}. Algorithm \ref{fedwb-frl} showcases the steps to train a global model under a heterogeneous federated reinforcement learning (HFRL) architecture. As we can see, it has a very similar structure to the idea of FedWB with the required semantics of reinforcement learning. After some number of training iterations we update the target network by equating it to the online network. We then perform an aggregation step in accordance with Algorithm \ref{model_fusion_algorithm}. As we perform the training, we will track the average amount of time each agent is capable of maintaining the pole upright before it fails and the episode ends. Furthermore, we will plot the average over 50 episodes to smooth out the result. Lastly, we will plot a comparison of the results using FedAvg (referred to as DQNAvg in \cite{pmlr-v151-jin22a}) versus FedWB in training a global model for HFRL. 

\begin{algorithm}[ht]
	\caption{FedWB for DQN with experience replay}
	\begin{algorithmic}[1]
	\vspace{.07cm}
        \REQUIRE Online Q-network: $\mathcal{Q}$
        \REQUIRE Target Q-network: $\hat{\mathcal{Q}}$
        \REQUIRE Experience replay object: $D$
        \REQUIRE $\epsilon$-greedy method
        \vspace{.07cm}
        \vspace{.07cm}
        
        \FOR{Episode: $1,\dots,E$}
        \FOR{$t=1,\dots,T$}
        \STATE Select an action according to an $\epsilon$-greedy strategy
        \STATE Update the state and environment
        \STATE Store the transition in $D$
        \STATE Sample $B$ from $D$
        \STATE Perform gradient step on squared error: $\big( \hat{\mathcal{Q}} - \mathcal{Q}\big)^2$ 
        \IF {C steps}
            \STATE Set $\hat{\mathcal{Q}} = \mathcal{Q}$ for all local models
            \STATE Perform aggregation process using model fusion with WBs (Algorithm \ref{model_fusion_algorithm})
        \ENDIF
        \ENDFOR
        \ENDFOR
        % Compute the barycenters
  \end{algorithmic}
	\label{fedwb-frl}
\end{algorithm}

\section{Results}\label{results}
\subsection{Distributed MNIST}
The goal of the experiment is showcase the validity of our approach in a distributed MNIST architecture. The experiment is setup and simulated as explained in the problem formulation (\ref{model_fusion_problem_formulation_section}) section. It is important to note that the results shown may vary under different conditions. Our code ran on a single computer equipped with an NVIDIA GeForce RTX 2070. Furthermore, due to different hardware specs available in running simulations, our speed comparison is extrapolated based on architecture type rather than wall-clock time. Although possibly negligible, the miss-handling of distributed computing can lead to large communication overhead. For example, if we consider too many agents in the network, we end up with a communications bottleneck for synchronization and local model parameter aggregation, in turn off-setting any parallelization speedup gained.

\begin{figure}[!ht]
    \centering
    \includegraphics[scale=0.65]{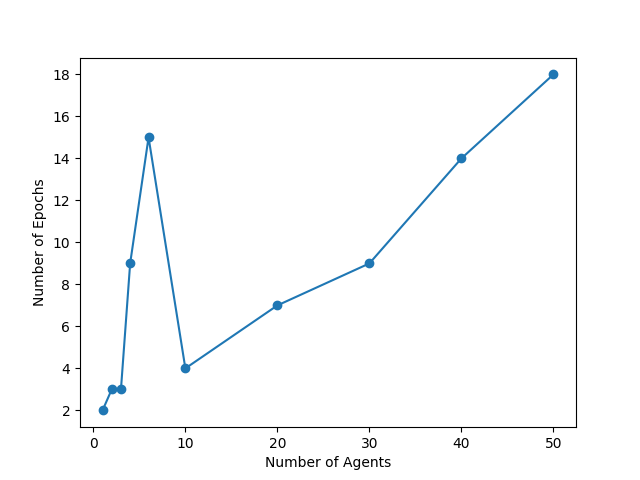}
    \caption{The number of epochs needed to reach at least 90\% accuracy given a number of agents in the distributed architecture.}
    \label{fig:epoch_per_agent_line_chart}
\end{figure}

The results of the experiment are elaborately explained in the following. Figure \ref{fig:epoch_per_agent_line_chart} demonstrates the number of epochs required to attain at least a 90\% accuracy of the global model given the free choice in the number of agents. The line graph shows an interesting, yet somewhat expected, positive relationship between the number of agents used and the number of epochs required. As we increase the number of agents in our network, the more epochs are required to reach certain degrees of testing accuracy. It is important to note that there is a fixed amount of data (60,000 images in training and 10,000 in testing) to create the network. Given the limited nature of the data, the positive relationship becomes self-explainable and self-evident. The interesting, not easily explained, behavior is the increase in the number of epochs from 4 to 5 agents, followed by the rapid decrease when we jump to 10 agents. This phenomenon was neither expected, nor will it be studied or explored in this paper. Our current conjecture is that the distribution of the data when splitting it into buckets hold some interesting properties that yield the difficulty in training convergence. For example, think of the sub-dataset of MNIST comprised of just 1s. There are a few variations on how people write the number 1. Some use only a vertical line while others use the bottom horizontal line and the wick on the top of the one. If we partition this dataset into two clearly different patterns, yields mutually exclusive sets, then training of local models may converge more easily. On the other hand, if we have equally distributed sets with equally distributed patterns, these local models may struggle to converge, although still being able to do so at some point. 
On another note, the convergence rate of a distributed architecture that has many agents is slow due to the contribution of the change of each agent since it is scaled by the number of agents in the network. Henceforth, another interesting property to consider is how the speed of convergence of the global model is affected inversely by the number of agents in the network (the more agents, the slower the convergence). Table \ref{table:pct_time_comparison} shows the relationship of the number of agents to the theoretical convergence speed to a minimum of 90\% accuracy on the test dataset; this table was generated and is complementary to Figure \ref{fig:epoch_per_agent_line_chart}. By theoretical we mean we do not account for the amount of wall-clock time required to maintain synchronization steps, as mentioned prior. To compute the values on the table, we take the wall-clock time taken to run the simulation for a single agent in the network and use this value to compute the speed to train over one epoch. Since training with one agent required two epochs, we multiply this value by two and use it as our base value to compare against. Then, we normalize the one-epoch time previously computed by dividing by the number of agents in the network; this yields the amount of time it takes for one epoch in the distributed architecture. Multiplying by the number of required epochs and dividing by the wall-clock time for the single agent simulation, we obtain the amount of time required to train the multi-agent architecture ($n$ agents) as a percentage of the time for one agent. As we can see, in generality increasing the agents results in less time to reach at least 90\% global accuracy on the test data, under the conditions described.
\begin{table}[!ht]
    \centering
    % \centering
    \caption{Amount of time (as \% of 1 agent) to reach a testing accuracy of at least 90\%,}
    \label{table:pct_time_comparison}
    % \begin{tabular}{|p{3.7cm}|p{3.7cm}|} 
    \begin{tabular}{cl}
        \toprule
        Number of Agents & Time\\
        \midrule
        1 & 1 \\
        2 & 0.75 \\
        3 & 0.50 \\
        4 & 1.125 \\
        5 & 1.25 \\
        10 & 0.20 \\
        20 & 0.175 \\
        30 & 0.15 \\
        40 & 0.175 \\
        50 & 0.18 \\
        \bottomrule
    \end{tabular}    
\end{table}

\begin{figure}[!ht]
    \centering
    \includegraphics[scale=0.65]{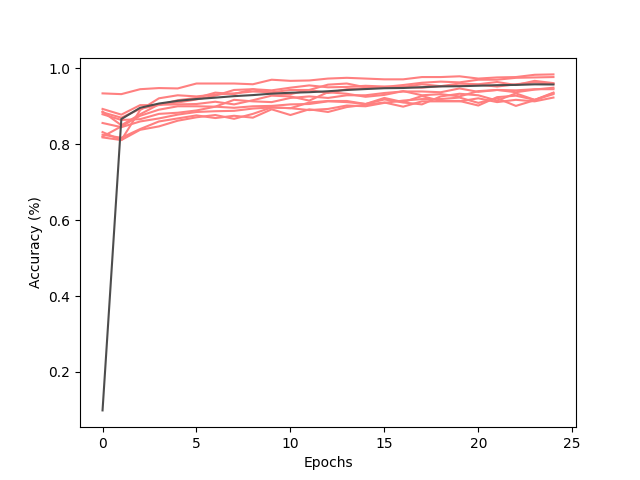}
    \caption{Accuracy per number of epochs. Results yielded from an architecture using 10 agents. The black line is the accuracy for the global model. The various red lines are the accuracy values for the various agents and their respective local models.}
    \label{fig:local_v_global_acc}
\end{figure}

Next, we focus on Figure \ref{fig:local_v_global_acc}, where we demonstrate the behavior of local models versus the global model. The graph demonstrates the results given that we choose to make our network with 10 agents. An interesting point one notices immediately is the jump from near 0\% accuracy to over 80\% accuracy for the global model, whereas the local models started at over 80\%. We believe this can be explained by local models taking locally optimal trajectories that would lead to the global minima. After making a single global aggregation step, the new starting point is non-optimal but lead the local models towards the global minima. As they train further, local models become more in sync as gradient steps shrink as they approach the optimal parameters. As we see, reaching nearly 25 epochs yields a global accuracy of above 95\% while local models span from 90\% to nearly 100\% accuracy.

\begin{figure}[!ht]
    \centering
    \includegraphics[scale=0.65]{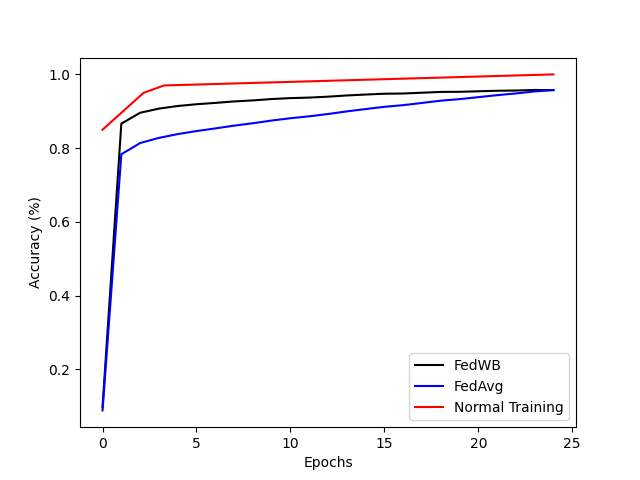}
    \caption{Model comparison between normally training with one agent (red line), FedWB (black line), and FedAvg (red line).}
    \label{fig:final_graph}
\end{figure}

Lastly, we look at Figure \ref{fig:final_graph}, where we have a final comparison of traditional training of a NN, the FedAvg algorithm, and our FedWB model. Starting with normal training, we observe it has a high starting point of above 80\% accuracy, eventually converging to 100\% accuracy on the test dataset. Next, we can directly compare FedWB with FedAvg. We notice that while they converge to the same results, their beginning trajectories are different. FedWB has a more rapid increase in test accuracy but flattens out quickly. On the other hand, FedAvg takes longer to reach more acceptable results, it does not flatten like FedWB. It is important to note that merely looking at epochs does not equate to speed as it takes varying amount of time to complete a singular epoch, as explained prior. Training a NN according to normal means has a higher final accuracy but it takes a longer amount of time to run 25 epochs than FedWB and FedAvg. On the same note, when comparing FedWB and FedAvg we have to take into account the speed of computing the Wasserstein barycenter versus a traditional average. The traditional average is a lighter computation that can be run much more quickly, but it is prone to more jumpy moves, as we see by the fact that it requires more epochs to reach high accuracy. On the other hand, using WB leads to a faster early convergence, but it requires more wall-clock time to run in comparison. Therefore, when choosing which algorithm to use, a practitioner must consider all these factors. It is possible a hybrid formulation yields the optimal solution, where for the first few epochs FedWB yields a better global model which is then swapped out for the speed of FedAvg, to reach potentially higher accuracy in a faster time frame.

\subsection{HFRL - CartPole}

\begin{figure}[!ht]
    \centering
    \includegraphics[scale=0.65]{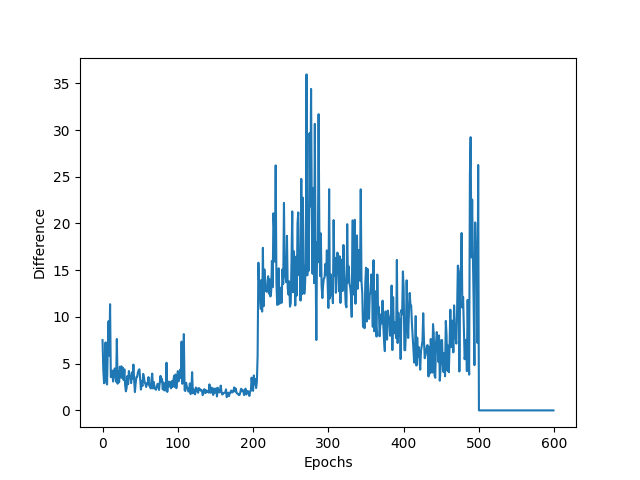}
    \caption{Difference in the "duration" per epoch for FedWB based and FedAvg based FHRL.}
    \label{fig:rl_diff}
\end{figure}

The metric we utilize to measure the success of the model is the amount of time it is able to balance the the pole. The goal of the experiment, similarly to the previous section, is to demonstrate the utility of FedWB in a new domain. Unlike training a neural network to make predictions over the MNIST dataset, the CartPole problem is significantly harder; equally so is the training of a DQN. Methods have been created to improve the stability of training DQNs, such as using a target network and an online network, where predictions are made towards the target network (the metaphorical oracle that gives us the value function we update the online network towards). Even with such a tool, the path towards improvement seems much more stochastic than the MNIST problem. In addition to improving the training stability, results are affected by various hyperparameters; most importantly, the annealing schedule and value of the $e$-greedy technique. Initially, more exploration means higher randomized "duration" of control as the model is trying out various actions to explore the space. As it progresses through training, the annealing schedule shrinks the probability of exploring the space, increases the likelihood we use the actions that yield the highest expected reward, as indicated by the DQN. Trained on a max of 600 episodes, the following figure (\ref{fig:rl_results_comparison}) shows the "duration" the model maintained the pole upright. First to notice is the temporal variance of the individual global models. Looking at the 50-epoch moving averages we see general trends of improvement of both global models. From figure \ref{fig:rl_results_comparison}, there are two main results to focus on. First is how FedWB begins with and sustains higher ability to balance the pole. The second main point of focus is how around the 450th epoch, both models are very close in capability. By the end of training epochs, we noticed both models we capable of balancing the pole to the maximum allowed number of actions, 500 actions. Limiting the maximum number of actions is common in  reinforcement learning to prevent an infinite loop. Furthermore, figure \ref{fig:rl_diff} shows the difference between FedWB and FedAvg on a per-epoch basis. To breakdown what is happening in the "difference" graph, we also need to look at the capabilities of the model as they trained over each epoch. We notice that for the first 200 epochs, there is a lot of exploration and thus poor control. There is then a change in control paradigm after the 200th epoch, where both models now become capable of balancing the pole for much longer. At this point, exploration is traded more often for exploitation (a lower $\epsilon$ value) and thus on average both models have higher control capability. Lastly, after the 450th epoch, there is another boom in the ability of the models to control the pole, where the models become capable of hitting the maximum-allowed number of actions (i.e., they've trained to a point where they are very good at balancing the pole). These three paradigms are easily seen in figure \ref{fig:rl_diff}, and the convergence of both models reaching a level of "expertise" where the difference of the duration of control reaches and stays at 0.

\begin{figure}[!ht]
    \centering
    \includegraphics[scale=0.65]{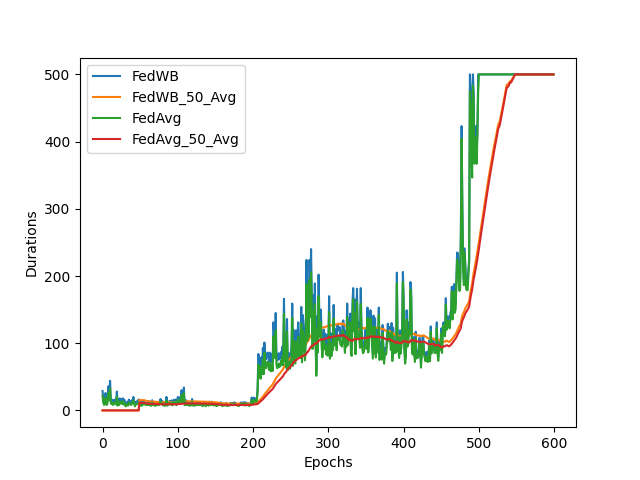}
    \caption{Result comparison between FedWB and FedAvg in training a global DQN for the CartPole problem.}
    \label{fig:rl_results_comparison}
\end{figure}

\section{Conclusions}\label{conclusion}
The primary goal of this paper was to bring the concepts of Wasserstein barycenter to new domains and to compare the novel algorithms we introduced with other models that are currently used in these domains. First, we began by introducing the concepts to build the FedWB algorithm, where we leverage WBs to generate a global neural network to make predictions over the MNIST dataset. We further compared it with building a global neural network using FedAvg. Although both models were capable of reaching very high accuracy on the training sets, there was a tradeoff in initial accuracy for speed. Because WB relies on solving the OT problem, it requires a significant amount of time to perform the aggregation step in comparison with arithmetic averaging; this time lost yield the gain in accuracy, as represented in results section. Furthermore, we expanded the FedWB architecture to federated heterogeneous reinforcement learning, where we trained a global model to perform control across various CartPole environments. Yet again we see the speed for accuracy trade off between FedWB and FedAvg, with the final steps having a convergence in model capabilities (either prediction accuracy for MNIST data, or control for RL). While FedWB leads in the early stages of training, FedAvg eventually catches up. The additional cost of training FedWB shows there is an important interplay between the methods when placed in juxtaposition. FedWB should be used initially to quickly yield better results, while eventually switching to FedAvg to reach a final global model quickly. The interplay described is a model-version of the well-known tradeoff between speed and accuracy. As Optimal Transport methods improve, yielding faster algorithms, the optimal switching point from FedWB to FedAvg may lean more towards only using FedWB, where the difference in speed between both methods may no longer need to be considered.

% \begin{table}
%  \caption{Sample table title}
%   \centering
%   \begin{tabular}{lll}
%     \toprule
%     \multicolumn{2}{c}{Part}                   \\
%     \cmidrule(r){1-2}
%     Name     & Description     & Size ($\mu$m) \\
%     \midrule
%     Dendrite & Input terminal  & $\sim$100     \\
%     Axon     & Output terminal & $\sim$10      \\
%     Soma     & Cell body       & up to $10^6$  \\
%     \bottomrule
%   \end{tabular}
%   \label{tab:table}
% \end{table}

\section*{Acknowledgments}
This work was partially supported by the Graduate Assistantships in Areas of National Need (GAANN) fellowship from the Department of Education grant P200A210087.

%Bibliography
\bibliographystyle{unsrt}  
\bibliography{references}

\end{document}